\documentclass[11pt]{article}
\usepackage{acl2015}
\usepackage{times}
\usepackage{url}
\usepackage{latexsym}
\usepackage{verbatim}
\usepackage{csquotes}
\usepackage{graphicx}
\usepackage{subcaption}
\usepackage[colorinlistoftodos]{todonotes}

\title{Evaluating Informal-Domain Word Representations With UrbanDictionary}

\author{Naomi Saphra \\
  University of Edinburgh\\
  {\tt n.saphra@ed.ac.uk}
\And Adam Lopez \\
  University of Edinburgh\\
  {\tt alopez@inf.ed.ac.uk}}

\date{}

\begin{document}
\maketitle
\begin{abstract}
Existing corpora for intrinsic evaluation are not targeted towards tasks in informal domains such as Twitter or news comment forums. We want to test whether a representation of informal words fulfills the promise of eliding explicit text normalization as a preprocessing step. One possible
evaluation metric for such domains is the proximity of spelling variants. We propose how such a metric might be computed and how a spelling variant dataset can be collected using UrbanDictionary.
\end{abstract}

\section{Introduction}
Recent years have seen a surge of interest in training effective models for informal domains such as Twitter or discussion forums. Several new works have thus targeted social media platforms by learning word representations specific to such domains \cite{tang2014learning}; \cite{benton2016learning}.

Traditional NLP techniques have often relied on text normalization methods when applied to informal domains. For example, ``u want 2 chill wit us 2nite'' may be transcribed as ``you want to chill with us tonight'', and the normalized transcription would be used as input for a text processing system. This method makes it easier to apply models that are successful on formal language to more informal language. However, there are several drawbacks to this method.

Building an accurate text normalization component for a text processing pipeline can require substantial engineering effort and collection of manually annotated training data. Even evaluating text normalization models is a difficult problem and often subjective \cite{eisenstein2013bad}.

Even when the model accurately transcribes informal spelling dialects to a standard dialect, text normalization methods may not be appropriate. Converting text to a style more consistent with The Wall Street Journal than Twitter may make parsing easier, but it loses much of the nuance in a persona deliberately adopted by the writer. Twitter users often express their spoken dialect through spelling, so regional and demographic information may also be lost in the process of text normalization \cite{eisenstein2013phonological}.

Distributional word representations hold promise to replace this flawed preprocessing step. By making the shared semantic content of spelling variants implicit in the representation of words, text processing models can be more flexible. They can extract persona or dialect information while handling the semantic or syntactic features of words \cite{benton2016learning}.

In this proposal, we will present a method of evaluating whether a particular set of word representations can make text normalization unnecessary. Because the intrinsic evaluation we present is inexpensive and simple, it can be easily used to validate representations during training. An evaluation dataset can be collected easily from UrbanDictionary by methods we will outline.

\section{Evaluating By Spelling Variants}

Several existing metrics for evaluating word representations assume that similar words will have similar representations in an ideal embedding space. A natural question is therefore whether a representation of words in social media text would place spelling variants of the same word close to each other. For example, while the representation of ``ur'' may appear close to ``babylon'' and ``mesopotamia'' in a formal domain like Wikipedia, on Twitter it should be closer to ``your''.

We can evaluate these representations based on the proximity of spelling variants. Given a corpus of common spelling variant pairs (one informal variant and one formal), we will accept or reject each word pair's relative placement in our dictionary. For example, we may consider \verb=(ur, your)= to be such a pair. To evaluate this pair, we rank the words in our vocabulary by cosine-similarity to \verb=ur=.

We could then count the pair correct if \verb=your= appears in the top $k$ most similar tokens. A similar method is common in assessing performance on analogical reasoning tasks \cite{mikolov2013distributed}. Having thus accepted or rejected the relationship for each pair, we can summarize our overall performance as accuracy statistic.

The disadvantage of this method is that performance will not be robust to vocabulary size. Adding more informal spelling variants of the same word may push the formal variant down the ranked list (for example, \verb=yr= may be closer to \verb=ur= than \verb=your= is). However, if these new variants are not in the formal vocabulary, they should not affect the ability to elide text normalization into the representation.

To make the metric robust to vocabulary size, instead of ranking all tokens by similarity to the first word in the variant pair, we rank only tokens that we consider to be formal. We consider a token to be formal if it appears on a list of formal vocabulary. Such a list can be collected, for example, by including all vocabulary appearing in Wikipedia or the Wall Street Journal.

\section{Gathering Spelling Variants}

If we have an informal text corpus, we can use it to generate a set of likely spelling variants to validate by hand. An existing unsupervised method to do so is outlined as part of the text normalization pipeline described by \cite{Gouws}. 

This technique requires a formal vocabulary corpus such as Wikipedia as well as a social media corpus such as Twitter. They start by exhaustively ranking all word pairs by their distributional similarity in both Wikipedia and Twitter. The word pairs that are distributionally similar in Twitter but not in Wikipedia are considered to be candidate spelling variants. These candidates are then re-ranked by lexical similarity,  providing a list of likely spelling variants. 

This method is inappropriate when collecting datasets for the purpose of evaluation. When we rely on co-occurrence information in a social media corpus to identify potential spelling variants, we provide an advantage to representations learned using co-occurrence information. When we rely on lexical similarity to find variants, we also offer an unfair advantage to representations that include character-level similarity as part of the model, such as \cite{dhingra}.

We therefore collected a dataset from an independent source of spelling variants, UrbanDictionary.

\subsection*{UrbanDictionary}

UrbanDictionary is a crowd-compiled dictionary of informal words and slang with over 7~million entries. We can use UrbanDictionary as a resource for identifying likely spelling variants. One advantage of this system is that UrbanDictionary will typically be independent of the corpus used for training, and therefore we will not use the same training features for evaluation.

To identify spelling variants on UrbanDictionary, we scrape all words and definitions from the site. In the definitions, we search for a number of common strings that signal spelling variants. To cast a very wide net, we could search for all instances of ``spelling'' and then validate a large number of results by hand. More reliably, we can search for strings like:

\begin{itemize}
\item misspelling of [\verb=your=]\footnote{Brackets indicate a link to another page of definitions, in this case for ``your''.}
\item misspelling of ``\verb=your=''
\item way of spelling [\verb=your=]
\item spelling for [\verb=your=]
\end{itemize}

A cursory filter will yield thousands of definitions that follow similar templates. The word pairs extracted from these definitions can then be validated by Mechanical Turk or study participants.

Scripts for scraping and filtering UrbanDictionary are released with this proposal, along with a small sample of hand-validated word pairs selected in this way\footnote{https://github.com/nsaphra/urbandic-scraper}.

\section{Experiments}

\begin{figure*}
\center
\verb=spelling[^\.,]* ('|\"|\[)(?P<variant>\w+)(\1)=
\caption{Regular expression to identify spelling variants.}
\label{re}
\end{figure*}

Restricting ourselves to entries for ASCII-only words, we identified 5289 definitions on UrbanDictionary that contained the string ``spelling''. Many entries explicitly describe a word as a spelling variant of a different ``correctly'' spelled word, as in the following definition of ``neice'':

\begin{displayquote}
Neice is a common misspelling of the word niece, meaning the daughter of one's brother or sister. The correct spelling is niece.
\end{displayquote}

Even this relatively wide net misses many definitions that identify a spelling variant, including this one for ``definately'':

\begin{displayquote}
The wrong way to spell definitely. 
\end{displayquote}

We extracted respelling candidates using the regular expression in Figure~\ref{re}, where the group \verb=variant= contains the candidate variant. We thus required the variant word to be either quoted or a link to a different word's page, in order to simplify the process of automatically extracting the informal-formal word pairs, as in the following definition of ``suxx'':

\begin{displayquote}
{}[Demoscene] spelling of "Sucks". 
\end{displayquote}

We excluded all definitions containing the word ``name'' and definitions of words that appeared less than 100 times in a 4-year sample of English tweets. This template yielded 923 candidate pairs. 7 of these pairs were people's names, and thus excluded. 760 (83\%) of the remaining candidate pairs were confirmed to be informal-to-formal spelling variant pairs.

Some definitions that yielded false spelling variants using this template, with the candidate highlighted, were:
\begin{enumerate}
\item recieve: The spelling bee champion of his 1st grade class above me neglected to correctly spell ``\emph{acquired}'', so it seems all of you who are reading this get a double-dose of spelling corrections.
\item Aryan: The ancient spelling of the word ``\emph{Iranian}''.
\item moran: The correct spelling of moran when posting to [\emph{fark}]
\item mosha: \ldots However, the younger generation (that were born after 1983) think it is a great word for someone who likes ``Nu Metal'' And go around calling people fake moshas (or as the spelling was originally ``\emph{Moshers}''. 
\end{enumerate}

Most of the false spelling variants were linked to commentary about usage, such as descriptions of the typical speaker (e.g., ``ironic'') or domains (e.g., ``YouTube'' or ``Fark'').

When using the word pairs to evaluate trained embeddings, we excluded examples where the second word in the pair was not on a formal vocabulary list (e.g., "Eonnie", a word borrowed from Korean meaning "big sister", was mapped to an alternative transcription, "unni").

\subsection{Filtering by a Formal Vocabulary List}

Some tokens which UrbanDictionary considers worth mapping to may not appear in the formal corpus. For example, UrbanDictionary considers the top definition of ``braj'' to be:

\begin{displayquote}
Pronounced how it is spelled. Means bro, or dude. Developed over numerous times of misspelling [brah] over texts and online chats.
\end{displayquote}

Both ``braj'' and ``brah'' are spelling variants of ``bro'', itself an abbreviation of ``brother''. If we extract \verb=(braj, brah)= as a potential spelling pair based on this definition, we cannot evaluate it if \verb=brah= does not appear in the formal corpus. Representations of these words should probably reflect their similarity, but using the method described in Section~2, we cannot evaluate spelling pairs of two informal words.

Using a vocabulary list compiled from English Wikipedia, we removed 140 (18\%) of the remaining pairs. Our final set of word pairs contained 620 examples.

\subsection{Results on GloVe}

As a test, we performed an evaluation on embeddings trained with GloVe \cite{pennington2014glove} on a 121GB English Twitter corpus. We used a formal vocabulary list based on English Wikipedia. We found that 146 (24\%) of the informal word representations from the word pairs in our dataset had the target formal word in the top 20 most similar formal words from the vocabulary. Only 70 (11\%)  of the informal word representations had the target formal word as the most similar formal word.

The word pairs with representations that appeared far apart often featured an informal word that appeared closer to words that were related by topic, but not similar in meaning. The representation of ``orgasim'' was closer to a number of medical terms, including ``abscess'', ``hysterectomy'', ``hematoma'', and ``cochlear'', than it was to ``orgasm''.

Other word pairs were penalized when the ``formal'' vocabulary list failed to filter out informal words that appeared in the same online dialect. The five closest ``formal'' words to ``qurl'' (``girl''), which were ``coot'', ``dht'', ``aaw'', ``luff'', and ``o.k''. 

Still other word pairs were counted as wrong, but were in fact polysemous. The representation of ``tarp'' did not appear close to ``trap'', which was its formal spelling according to UrbanDictionary. Instead, the closest formal word was ``tarpaulin'', which is commonly abbreviated as ``tarp''.

These results suggest that current systems based exclusively on distributional similarity may be insufficient for the task of representing informal-domain words. 

\section{Biases and Drawbacks}

Evaluating performance on spelling variant pairs could predict performance on a number of tasks that are typically solved with a text normalization step in the system pipeline. In a task like sentiment analysis, however, the denotation of the word is not the only source of information. For example, a writer may use more casual spelling to convey sarcasm:

\begin{displayquote}
I see women who support Trump or Brock Turner and I'm like ``wow u r such a good example for ur daughter lol not poor bitch'' (Twitter, 18 Jun 2016)
\end{displayquote}

or whimsy:

\begin{displayquote}
*taking a personalitey test*\\
ugh i knew i shoud have studied harder for this (Twitter, 6 Jun 2016)
\end{displayquote}

An intrinsic measure of spelling variant similarity will not address these aspects.




Some of the disadvantages of metrics based on cosine similarity, as discussed in \newcite{DBLP:journals/corr/FaruquiTRD16}, apply here as well. In particular, we do not know if performance would correlate well with extrinsic metrics; we do not account for the role of word frequency in cosine similarity; and we cannot handle polysemy. Novel issues of polysemy also emerge in cases such as ``tarp''; ``wit'', which represents either cleverness or a spelling variant of ``with''; and ``ur'', which maps to both ``your'' and ``you are''.

However, compared to similarity scores in general \cite{gladkova},  spelling variant pairs are less subjective.

\section{Conclusions}

The heuristics used to collect the small dataset released with this paper were restrictive. It is possible to collect more spelling variant pairs by choosing more common patterns (such as the over 5000 entries containing the string ``spelling'') to pick candidate definitions. We could then use more complex rules, a learned model, or human participants to extract the spelling variants from the definitions. However, the simplicity of our system, which requires minimal human labor, makes it a practical option for evaluating specialized word embeddings for social media text.

Our experiments with GloVe indicate that models based only on the distributional similarity of words may be limited in their ability to represent the semantics of online speech. Some recent work has learned representations of embeddings for Twitter using character sequences as well as distributional information \cite{dhingra}; \cite{vosoughi}. These models should have a significant advantage in any metric relying on spelling variants, which are likely to exhibit character-level similarity.

\bibliographystyle{acl}
\bibliography{repeval}

\end{document}